\documentclass{llncs}
\usepackage{times,graphicx,hyperref}

\begin{document}

\title{The DeLiVerMATH project}
\subtitle{Text analysis in mathematics}
\author{Ulf  Sch\"{o}neberg\inst{} \and  Wolfram Sperber\inst{}}
\authorrunning{Ulf Sch\"{o}neberg, Wolfram Sperber}
\tocauthor{Ulf Sch\"{o}neberg and Wolfram Sperber}
\institute{FIZ Karlsruhe/Zentralblatt MATH, Franklinstr. 11, 10587 Berlin, Germany}

\maketitle

\begin{abstract}
A high-quality content analysis is essential for retrieval functionalities but the manual extraction of key phrases and classification is expensive. Natural language processing provides a framework to automatize the process. Here, a machine-based approach for the content analysis of mathematical texts is described. A prototype for key phrase extraction and classification of mathematical texts is presented. 
%\keywords{content analysis, key phrase extraction, classification, natural language processing} 
\end{abstract}

\section{Introduction}
The database zbMATH~\cite{zbMATH} provided by FIZ Karlsruhe/Zentralblatt MATH is the most comprehensive bibliographic reviewing service in  mathematics. Both key phrases and classification of the mathematical publications are central features of content analysis in zbMATH. Up to now, these data are created by expert which means  time and labor-expensive work.
 
In the last years, computational linguistics has developed concepts for natural language processing by combining linguistic analysis and statistics. These concepts and tools were used as a platform for our activities to develop machine-based methods for key phrase extraction and classification according to the Mathematical Subject Classification (MSC2010)~\cite{MSC}.

The DeLiVerMATH project funded by the Deutsche Forschungsgemeinschaft is a common activity of the library TIB Hannover, the research center L3S Hannover and FIZ Karlsruhe. It started in March 2012.

\section{The prototype}

We are starting with a presentation of a prototype extracting key 
phrases and classifying a mathematical text. Snapshot on 
Figure~1 demonstrates its functionality. 

\begin{figure}[hbt]
    \centering
    \includegraphics[width=\textwidth]{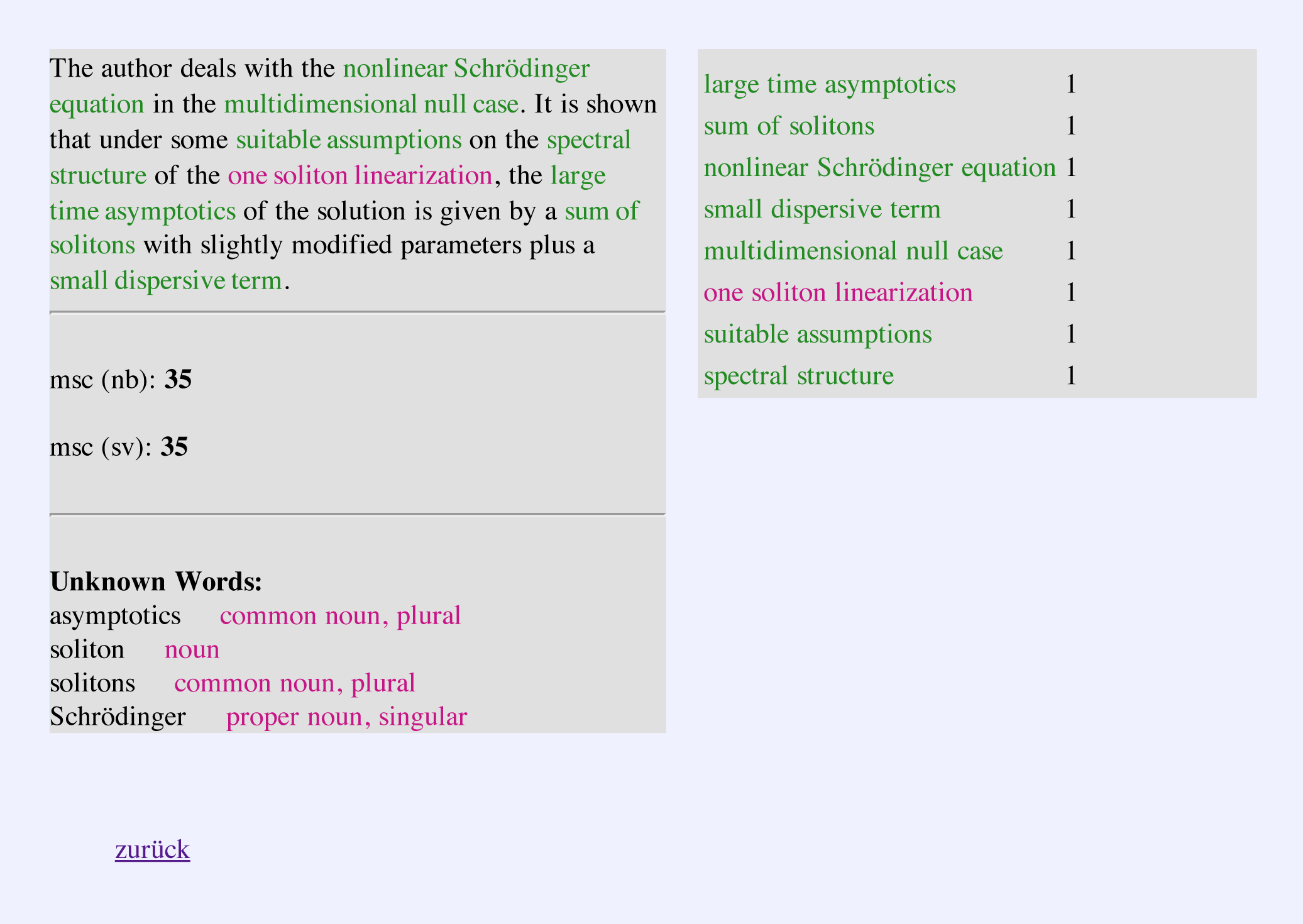}
    \caption{The user interface of the prototype}
    \label{userinterface}
\end{figure}

\noindent The original text, here a review from zbMATH, is located in the left box on the top. The input can be -- in principle -- an arbitrary mathematical text.

The extracted candidates for key phrases and their frequencies are presented in the list on the right side and are also highlighted in the original text.
 
The proposed MSC classes calculated with Naive Bayes (nb) and Support Vector Machines (sv) are shown below the input text. Currently, the classification is restricted to the top level of the MSC.

Moreover, a list of unknown tokens (tokens outside of dictionary) together with a proposed word class is given.

\section{Natural Language Processing (NLP) in mathematical publications}
Existing Open Source tools and dictionaries from NLP were adapted to the special needs of mathematical texts.  
NLP provides a broad spectrum of methods for text analysis, especially
\begin{itemize}
\item Segmentation to identify text units 
\item Tokenization, the process of breaking a text stream into words, symbols and formulae, or other
meaningful elements called tokens
\item Morphological Parsing for a linguistic analysis of tokens 
\item Part-Of-Speech (PoS) tagging, the classification of a token
within a text, e.g., `convergence' as a noun. 
\item Parse Tree, the identification of associated text fragments, e.g., of a noun phrase 
\item Named Entity Recognition, the detection of phrases typically used by a community
\end{itemize}

\noindent Especially, the PoS tagging is of fundamental importance for the text analysis. PoS tagging requires a classification scheme for the tokens. Here we use the Penn Treebank PoS scheme~\cite{PoS} consisting of 45 tags for words and punctuation symbols. This scheme has relevant drawback for mathematical texts:  no special tag for mathematical formulae. In our approach, mathematical formulae will be handled  with an auxiliary construct:  formulae (which are available as TeX code) are transformed to special nouns.  This allows us to extract phrases which contain formulae beside the English text.  A more detailed analysis of the mathematical formulae will be done in the MathSearch project of FIZ Karlsruhe and Jacobs University of Bremen.   
The PoS tagging use large dictionaries to assign a tag to a token. Here, the Brown corpus~\cite{BrownCorpus}  is used as a dictionary covering  more than 1,000,000 English words which are classified according to the Penn Treebank scheme. 
Some NLP tools are provided as Open Source software. Within the DeLiVerMATH project, the Stanford PoS Tagger~\cite{StanfordTag} is used.

There are two problems: the ambiguity of PoS tags (many tokens of the corpus can belong to more than one word class) and unknown words (mathematics has a lot of tokens outside the Brown corpus). For both problems, a suitable PoS tag of a token in a phrase can be determined using the Viterbi algorithm, a dynamic programming technique. Moreover, mathematics relevant dictionaries are under development 
\begin{description}
\item[Acronyms]
Often special spellings are used for acronyms (capitals) which can be easily identified in texts by heuristic methods. A  database of 3,000 acronyms with their possible resolutions was build up and implemented. 
\item[Mathematicians]
The author database of zbMATH covers the names of more than 840,000 mathematicians which can be used to identify names of mathematicians in the text phrases.
\item[Named Mathematical Entities]
Named mathematical entities are phrases which are used for a special mathematical object by the mathematical community. It is planned to integrate existing lists of named mathematical entities and other vocabularies, 
especially the Encyclopedia of Mathematics~\cite{EoM}, PlanetMath~\cite{PlanetMath}, and the vocabularies of the mathematical part of Wikipedia and the MSC.
\end{description}    
The analysis of single tokens is only the first step for the linguistic analysis. Also in mathematics, phrases are often much more important for the content analysis  than single tokens. Key phrase extraction requires the identification of chunks. Noun phrases are the most important candidates for key phrases. The first version of our tool searches for distinguished PoS tag patterns of noun phrases in the zbMATH items (abstracts or reviews). Some problems  may arise, e.g., longer key phrases are extracted only partially. A more general approach is to use, instead of that, syntactic parsers for the identification of key phrases being based on grammars for the English language. A syntactic parser will be implemented into the prototype as the next step.
Not all extracted noun phrases are meaningful for the content analysis as the example shows. Here we plan to build up filters (dictionaries of often used irrelevant noun phrases).      
For the classification, different approaches and classifiers already 
are implemented on different corpora: abstract and reviews, key phrases, 
and also full texts (in planning). Up to now, the classification is 
restricted to the top level of the MSC, a finer classification is under way.    

\section {Evaluation}
The prototype will be evaluated by the editorial staff of Zentralblatt MATH to assign key phrases and classification to mathematical publications. The additional expense to evaluate the key phrases and classification  is low because the tool can be integrated and used in the daily workflow. The editors can reject a proposed key phrase and classification in an easy way.  We state that the number of key phrases seems to be significantly higher compared with the number of manually created key phrases which could be used for a better retrieval (e.g., to find similar publications) and the ranking. Some techniques to reject irrelevant phrases and detect similar phrases must be developed.  
For the classification, different methods are under development, the prototype shows the classification calculated with the Naive Bayes and the Support Vector Machine approach. It seems that the quality for the top level of the MSC is sufficient. But, we need a finer MSC classification in the future. We will also analyze the influence on the corpora to the classification and seek an answer to the question: Is it better to use key phrases instead of reviews and abstracts    or fulltexts for the classifiers?   
\section {Outlook}
The methods and tools are developed and checked for database zbMATH but can also be used for  the content analysis of full texts.
Up to now, the method is restricted to English texts. An extension to other languages is possible by adding  dictionaries and grammars of other languages.
A production version of the tool could be provided on the Web site of the FIZ Karlsruhe/\hskip0pt Zentralblatt Math as an additional service for the mathematical community. 
One objective of the project is to build up a controlled vocabulary of mathematics and to match this with the MSC. Therefore, the most frequent key phrases for the MSC classes will be determined. These phrases could be the base of a controlled vocabulary for mathematics, a helpful tool for the standardization of mathematical language and communication and a starting point for developing a thesaurus for mathematics.
The prototype is a first step towards a machine-based content analysis of mathematical publications. We are optimistic that the approach described above can contribute to a less expensive content analysis and allows an improved retrieval.

\end{document}